\documentclass[sigconf]{acmart}
\AtBeginDocument{%
  \providecommand\BibTeX{{%
    \normalfont B\kern-0.5em{\scshape i\kern-0.25em b}\kern-0.8em\TeX}}}

\setcopyright{acmcopyright}
\copyrightyear{2018}
\acmYear{2018}
\acmDOI{10.1145/1122445.1122456}

\providecommand{\mypara}[1]{\smallskip\noindent\emph{#1}}
\providecommand{\myparab}[1]{\smallskip\noindent\textbf{#1}}

\newcommand{\cale}{\mathcal{E}}

\newcommand{\calv}{\mathcal{V}}

\newcommand{\mr}{\mathbf{r}}
\newcommand{\mx}{\mathbf{x}}

\newcommand{\mX}{\mathbf{X}}

\usepackage{lipsum}
\acmJournal{TOG}
\acmVolume{37}
\acmNumber{4}
\acmArticle{111}
\acmMonth{8}

\usepackage{subfigure}
\usepackage{booktabs}
\usepackage{amsfonts} 
\usepackage{algorithm}
\usepackage{algorithmic}
\usepackage{mathtools}
\usepackage{xspace}
\usepackage[font=small,labelfont=bf]{caption}

\usepackage{graphicx}
\usepackage{colortbl}
\newcommand{\transpose}{\top}
\newcommand{\reals}{\mathbf{R}}

\newcommand{\mE}{\mathbf{E}}
\usepackage{natbib}
\providecommand{\mypara}[1]{\smallskip\noindent\emph{#1}}
\providecommand{\myparab}[1]{\smallskip\noindent\textbf{#1}}

\setcitestyle{numbers,citesep={,}}
 
\newcommand{\qq}[1]{\textcolor{blue}{#1}}

\usepackage{tikz}

\newcommand*\circled[1]{\protect\tikz [baseline=(char.base)]{
  \protect\node[shape=circle,draw,fill=black,text=white,font=\bf,inner sep=0.5pt] (char)
  {\scriptsize#1}; 
}} 
\makeatletter 
\let\@authorsaddresses\@empty 
\makeatother 
\setcopyright{none}  

\definecolor{colour3}{RGB}{178,55,250} 

\newcounter{noteMCctr} 

\setcopyright{none}
\renewcommand\footnotetextcopyrightpermission[1]{}
\newcommand{\modelname}{Equity2Vec\xspace}

\begin{document}
\title{\textsc{Equity2Vec}: End-to-end 
Deep Learning Framework for Cross-sectional Asset Pricing  
}

%

\author{Qiong Wu}
\affiliation{%
  \institution{William \& Mary}
  \streetaddress{}
  \city{}
  \country{}}
\email{qwu05@email.wm.edu}

\author{Christopher G. Brinton}
\affiliation{%
  \institution{Purdue University}
  \city{}
  \country{}}
  \email{cgb@purdue.edu}

\author{Zheng Zhang}
\affiliation{%
  \institution{William \& Mary}
  \streetaddress{}
  \city{}
  \country{}}
\email{zzhang14@email.wm.edu}

\author{Andrea Pizzoferrato}
\affiliation{%
  \institution{University of Bath \\ The Alan Turing Institute}
  \streetaddress{}
  \city{}
  \state{}
  \country{}}
  \email{ap2873@bath.ac.uk}

\author{Zhenming Liu}
\affiliation{%
  \institution{William \& Mary}
  \streetaddress{}
  \city{}
  \state{}
  \country{}
  \postcode{}}
\email{zliu20@wm.edu}

\author{Mihai Cucuringu}
\affiliation{%
 \institution{University of Oxford\\ The Alan Turing Institute}
 \city{}
 \state{}
 \country{}}
 \email{mihai.cucuringu@stats.ox.ac.uk}


\begin{abstract}
Pricing assets has attracted significant attention from the financial technology community. We observe that the existing solutions overlook the cross-sectional effects and not fully leveraged the heterogeneous data sets,
leading to sub-optimal performance. 


To this end, we propose an end-to-end deep learning framework to price the assets.  Our framework possesses two main properties: 1)~We propose \textsc{\modelname}, a graph-based component that effectively captures both long-term and evolving cross-sectional interactions. 2)~The framework simultaneously leverages all the available heterogeneous alpha sources including technical indicators, financial news signals, and cross-sectional signals. Experimental results on datasets from the real-world stock market show that our approach outperforms the existing state-of-the-art approaches. Furthermore, market trading simulations demonstrate that our framework monetizes the signals effectively.

%
%

%

%
%

\end{abstract}

\settopmatter{printacmref=false}

 \maketitle 
\pagestyle{plain}
\section{Introduction}
\vspace{2mm}

It is widely acknowledged  that forecasting stock prices is a difficult task.    
Most traditional efforts rely on time series analysis models, such as Autoregressive models~\cite{li2016stock}, Kalman Filters~\cite{musoff2009fundamentals}, and technical analysis~\cite{kirkpatrick2010technical,neely1997technical,neely2011technical}.  Deep neural networks, especially recurrent neural networks
(RNN)~\cite{hu2018listening,ticknor2013bayesian,ding2015deep,goccken2016integrating,zhang2017stock,rather2015recurrent,de2018advances} recently emerged as an effective solution for stock prediction tasks. Such lines of work have significantly increased in popularity in recent years, mostly fueled by the fact that a large collection of high-quality financial data sets have become available.

We observe two fundamental limitations in the prior works: 
\noindent{\emph{1. Cross-sectional effects are not properly leveraged.}}
Most existing approaches treat each stock independently and overlook the cross-sectional effect.  The cross-sectional effect posits the fact that the information from one stock may
influence/impact another stock's price change in both static and dynamic aspects. Statically, the stocks that share the same intrinsic properties may move synchronously. For instance, when Twitter goes up, Facebook is more likely to go up because they are in the same sector (social media advertising).  
Dynamically, stock relation is also temporally evolving. For example, in early 2021, AMC theatres,  GameStop, and BlackBerry suddenly exhibit co-movement driven by investors on social media who are buying up these stocks.



\noindent{\emph{2. Heterogeneous data sets are not leveraged to their fullest extent.}} Most models use only one type of data (i.e., either textual information or ``technical factors'' \cite{ming2014stock,hu2018listening,mittermayer2006newscats,wu2019adaptive,patel2015predicting}  in numeric form). The latter is derived from prices and traded volumes. It remains open to building a model that leverages heterogeneous data sources. This model needs to reconcile and aggregate information from different data sources.  


\begin{figure}[]
    \centering
    \subfigure[The co-mentions capture sector relation between stocks]{
    \includegraphics[scale = 0.19]{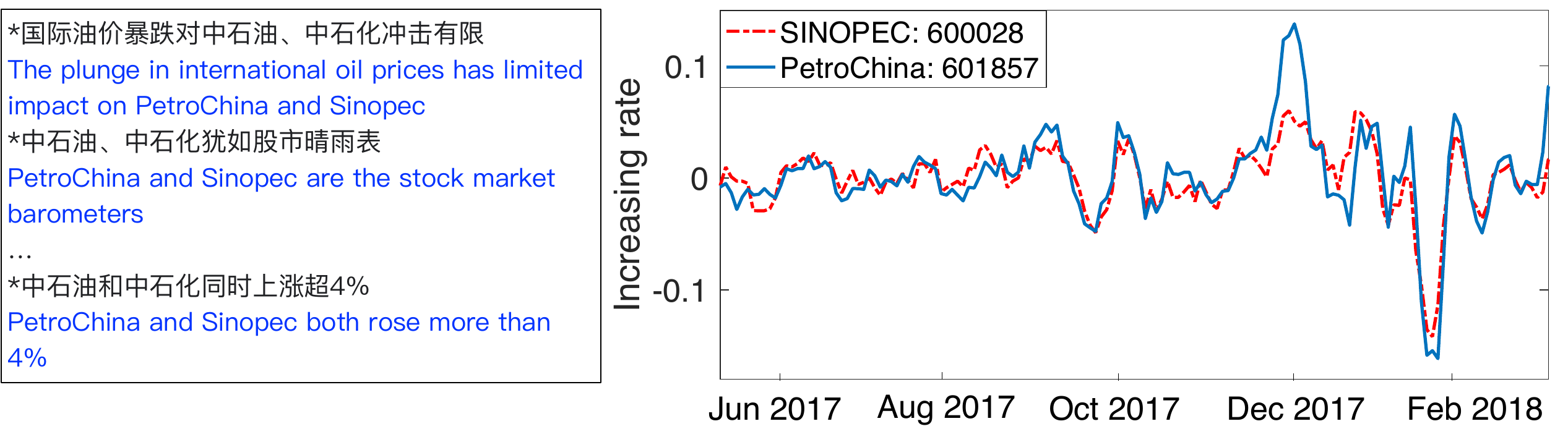}\label{fig:sec}
    }\hspace{0.05\textwidth}
    \subfigure[The co-mentions capture supply-chain relation between stocks]{
    \includegraphics[scale = 0.185]{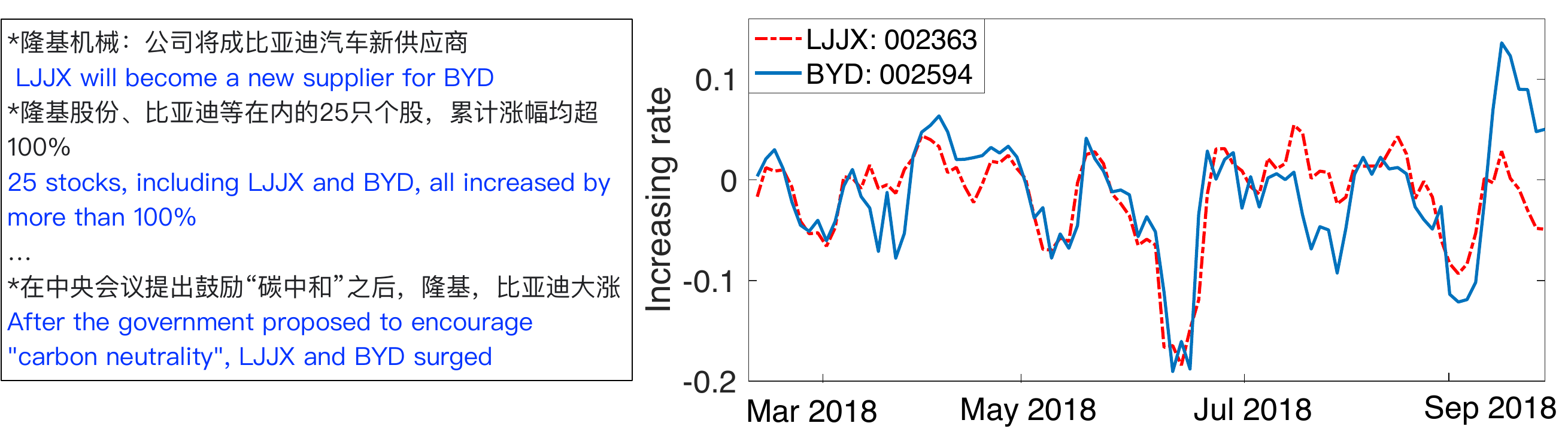}\label{fig:supply}}
    \vspace{-5mm}
    \caption{Examples showing our key observations: When the news mention stocks frequently, the stocks are 1) likely to reflect relations, such as sector and supply-chain, 2) likely to have similar movement on prices.}
    \vspace{-7mm}
    \label{fig:intro}
\end{figure}

An efficient stock embedding scheme that addresses the first limitation must determine: 1)~what data source includes the co-movement information, 2)~how to extract the cross-sectional signals from the data source, and 3)~how to
incorporate both the static and dynamic stock relations into the stock embedding. We propose \textsc{\modelname} that answers the three questions. 


Ideally, the stock representation should reflect comprehensive relations such as sector, supply chain, value, growth, business cycles, volatility, and analysts' confidence towards the stock. 
One possible way is to collect such information manually from experts and analysts, but it is inefficient and costly. Further, there are no widely agreed upon standard approaches to converts people's opinions into stock representations. \emph{Since millions of investors, analysts, and financial experts share opinions, events, comments, and transactions about stocks in the news, we consider using news as a data source to learn stock representations. }

Next, we make two key observations by analyzing news on stocks. When two stocks are frequently co-mentioned, 1)~they are likely to share common characteristics such as sector and supply-chain relation, 2)~their prices tend to have a similar trend. 
For example, the co-mentioned stocks in Figure~\ref{fig:intro} (a) are in the same sector (energy), while Figure~\ref{fig:intro} (b) shows they have a supply-chain relationship (i.e., LJJX is a supplier of BYD). In both cases, the prices of these co-mentioned stocks often move synchronously and most often in the same direction. \emph{Based on our observations, we use news co-mention\footnote{Through the statistic analysis of real-world online media, a piece of news refers to $3.0$ stocks on average. } to learn the stock representation. }


Moreover, \textsc{\modelname} extracts the long-term (static) and evolving (dynamic) stock relations by the following approach. To capture the long-term relation, we build a global stock co-occurrence matrix ~\cite{pennington2014glove} (see Figure~\ref{mf} (\circled{1})) with a long observation window. We extract the stocks' long-term representations via matrix factorization of the co-occurrence matrix. To learn the evolving relation, we build a stock graph that reflects the dynamic neighboring relations, where  \textsc{\modelname} propagates the embedding of a stock to its neighbors to capture the cross-sectional information of the stocks effectively.

To leverage the heterogeneous data sources (i.e., second limitation of prior works), we propose a framework that integrates the learned stock embeddings, news signals, and technical factors into a neural network model to make the final prediction.  
We perform extensive experiments on real-world data that contains more than 3,600 stocks in the Chinese stock market from 2009 to 2018. 
The experimental results demonstrate the effectiveness of stock representations extracted by \textsc{\modelname} outperforms existing state-of-the-art works. The market trading simulation illustrates that \textsc{\modelname} along with the proposed framework increases profit significantly. 

In summary, this paper makes the following contributions: 
\begin{itemize}
    \item We propose \textsc{\modelname} that incorporates news into stock embeddings. To the best of our knowledge, \textsc{\modelname} is the first work that mines the stock representation from the news co-mention. Moreover, \textsc{\modelname} captures both long-term (static) and evolving (dynamic) relations between stocks. 
    \item We forecast stock prices using multiple categories of signals (i.e., heterogeneous data sets), including cross-sectional embeddings, technical signals, and financial news signals. 
    \item Extensive experiments on real-world data sets confirm the efficacy of our approach, comparing favorably to state-of-the-art methods.  
\end{itemize}

%
%
%



\begin{figure*}[t!]
    \centering     
    \includegraphics[scale = 0.35]{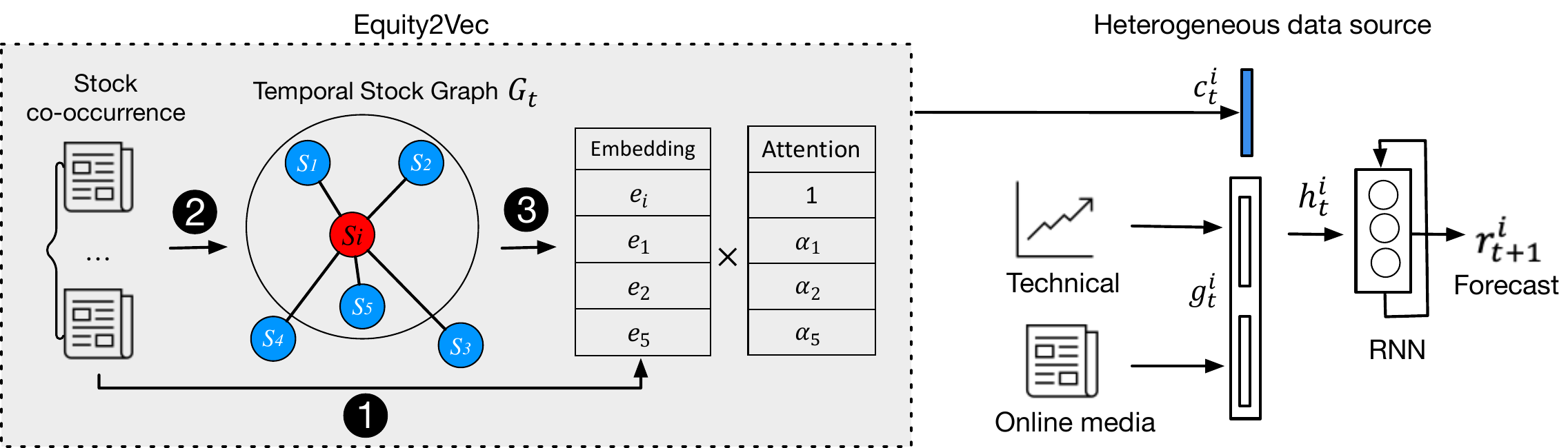}
    \caption{The illustration of our end-to-end framework. It contains the \textsc{\modelname} component (Section~\ref{sec_stock2vec}) and heterogeneous data source component (Section~\ref{data_source}).}
    \label{fig:framework}
    \vspace{-1mm}
\end{figure*}


\vspace{-2mm}
\section{Preliminaries and Framework overview}\label{pre}
\vspace{-1mm}
\myparab{Problem setting.}
Given a universe of $n$ stocks $s_1, s_2,...,s_n$, the stock price trend is log return for a given stock $i$ on day $t$
$r_{t}^{i} = \log \left(\frac{p_t^i}{p_{t-1}^{i}},  \right) $
where $p_t^i$ denotes the open price of stock $i$ on day $t$. We formulate the task of predicting the  future price trend as a regression problem. The response is the future return $r_{t+1}^i$, and  $\mx_{t}^i$ denotes the vector of features associated with stock $i$ on day $t$.  The historical features up to time $t$ are defined as $\mx_{\leq t}^i$.
We aim to learn the function $r_{t+1}^i = f(\mx_{\leq t}^i)$.

\vspace{-1mm}
\myparab{Framework Overview.} Figure~\ref{fig:framework} shows our overall   framework, which includes the \textsc{\modelname}  component and the heterogeneous data source component. The \textsc{\modelname} component first learns the stocks' long-term relations from the global stock co-occurrence matrix, and then extracts the static stock embedding as $e_i$~(\circled{1}).  Then, at time $t$, we build the temporal graph $G_t$ dynamically based on the local stock co-occurrence matrix to capture the evolving relations (\circled{2}). Within graph $G_t$, each solid circle represents a stock. Stocks close to one another are likely to be associated with a similar moving trend. Finally, our approach obtains the final stock representation $c_t^i$ by propagating its neighbors' basic embedding via an attention mechanism~(\circled{3}). 

In the heterogeneous data source component, we integrate the stock embedding ($c_t^i$) with dynamic input ($g_t^i$) (generated from technical factors and online textual data), and denote the heterogeneous output as $h_t^i$. Finally, the RNN model forecasts future return $r_{t+1}^i$ based on the input $h_t^i$.



\vspace{-2mm}
\section{\textsc{\modelname} from news}\label{sec_stock2vec}
\vspace{-1mm}


We propose \textsc{\modelname}, which mines the stock embeddings from the news, since such data comprises a valuable knowledge repository with rich relation information between stocks from the crowd of financial experts/journalists. Our approach is inspired by the observation (as depicted in Figure~\ref{fig:intro}) that stocks frequently co-mentioned by the same news are likely to share similar properties and exhibit co-movements in their price trends.  We formulate the stock co-occurrence matrix, and use matrix factorization to extract the \emph{static} stock representation in Section~\ref{static}. To explicitly leverage the cross-sectional signals and circumvent the challenge that the relations between stocks are evolving, we build a temporal stock graph and fine-tune the stock representations by \emph{dynamically} infusing neighbors latent representations (Section~\ref{dynamic}).

\vspace{-2mm}
\subsection{Capturing long-term stock relations}\label{static}
\vspace{-1mm}
We now focus on learning the long-term relation, and   discuss how to address the dynamic relations in the next subsection.

\begin{figure}[]
    \centering
    \subfigure[]{
    \includegraphics[width=0.26\textwidth]{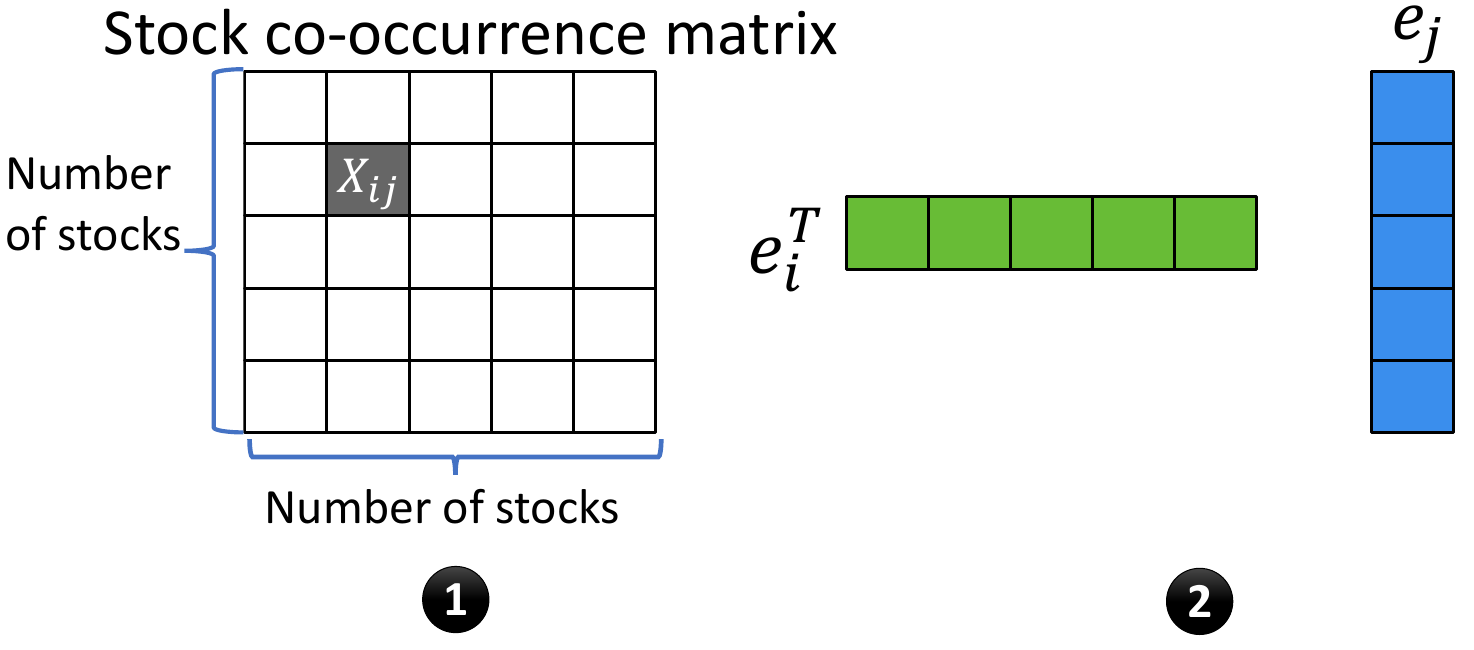}
    \label{mf}}  \hspace{0.01\textwidth}
    \subfigure[]{
    \includegraphics[width=0.18\textwidth]{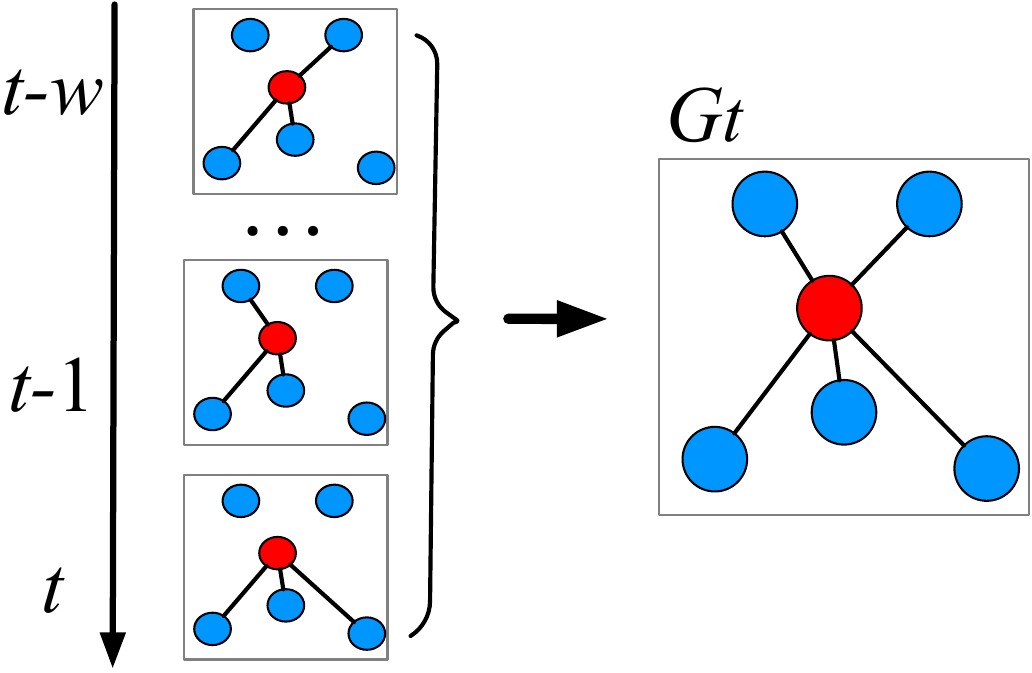}
    \label{temporal} }   
    \vspace{-6mm}
    \caption{(a) Our approach to build the stock co-occurrence matrix and calculate the static embedding for stocks. (b)~The construction of temporal graph.}
    \label{fig:main_chart}
    \vspace{-4mm}
\end{figure}

\vspace{-1mm}
\myparab{Co-occurrence Matrix.} 
We build a stock co-occurrence matrix by counting the stock co-occurrence within each news. We prefer the stock co-occurrence matrix instead of the entire news and stock matrix for the following two reasons: (1) The size of the stock co-occurrence matrix only depends on the number of stocks, and is much smaller than the news and stock matrix. 
(2) We only care about the stock representation, which thus renders news representations as not necessary. We use the global co-occurrence matrix to obtain static representations that reflect the essential relations between stocks in the long term. 


Formally, suppose we have $n$ stocks and $\mX \in \reals^{n\times n}$ denotes the stock co-occurrence matrix. Figure~\ref{mf} (\circled{1}) shows $\mX_{i, j}$, counting the number of articles that mention both $s_i$ and $s_j$ before testing phase. 
We associate a vector $e_i \in \reals^d$ to represent the stock $i$'s static representation, where $d$ is the dimension of the stock representation. In this way, the similarity between stock $i$ and stock $j$ can be formulated as the inner product of $e_i$ and $e_j$, given by  $e_i^{\transpose}e_j$.

\myparab{Matrix Factorization. }  Figure~\ref{mf} (\circled{2}) shows that we adopt matrix factorization~\cite{koren2009matrix,wang2021heals} to learn the embedding of the stocks. 
Here, matrix factorization works as a collaborative filtering method. 
The idea behind matrix factorization is to learn the latent representation where stocks near each other will likely obtain similar embeddings. 
Given the co-occurrence matrix reflects the similarity between stocks, we obtain the latent representation of stocks through fitting the training data by optimizing the objective function
\begin{equation}
J_s = \sum_{i,j=1}^{n}(e_i^{\transpose} e_j  - \mX_{ij})^2. 
\label{loss_raw}
\end{equation}

In reality, there could exist prior bias of stocks as the prior preference of financial journalists/experts. Hence,
we use $b_i$ and $b_j$ as the bias of
stock $i$ and stock $j$ and introduce them to the objective function,

\vspace{-2mm}
\begin{equation}
J_s = \sum_{i,j=1}^{n}(e_i^{\transpose} e_j  + b_i + b_j - \mX_{ij})^2 + \beta (||\theta||^2)
\label{loss_raw_regu}
\end{equation}

\noindent where  $||\theta|| = (||e_i||^2 + ||e_j||^2) + b_i^2 + b_j^2 $ and $\beta (||\theta||^2)$ is the regularization term that prevents overfitting.


\vspace{-2mm}
\subsection{Capturing evolving stock relations}\label{dynamic}
\vspace{-1mm}
The latent representation learned from the above global occurrence matrix reflects the static stock relations. Motivated by the fact that the stock relations are changing over time and cross-asset signals are beneficial towards stock price prediction, we further fine-tune the stock representation by building a temporal stock graph and infusing the neighbors' embedding dynamically.

\myparab{Temporal stock graph. }
As shown in Figure~\ref{temporal}, the construction of  $G_t$ consists of two steps.
(1) Construct the stock graph using the  co-occurrence matrix. Assume $\tilde{G_t} = \{\calv, \cale_t\}$ is the stock graph at time $t$, where $\calv = \{s_1, \dots, s_n\}$ is the set of stocks. $(s_i, s_j) \in \cale_t$ if and only if $s_i$ and  $s_j$ are co-mentioned by the news collected at time $t$. The edge eight over $s_i$ and $s_j$ is the number of 
co-occurrences across all news on date $t$.
(2) Due to insufficient number of news about specific stocks in time $t$, we maintain a sliding window $w$ ($w$ is a hyper-parameter) to collect a sequence of stock graphs and then construct $G_t$ by taking an exponential moving average of $\tilde{G}_{t-w},...,\tilde{G}_{t},\tilde{G}_t$, where we assign a nearby graph a larger weight.

\myparab{Propagation of neighbors' embedding via a stock attention mechanism.}
Given the temporal graph ($G_t$) identifies the dynamic stock structure, it is crucial to appropriately update the stock representation by infusing the current neighbors' embedding. 
Motivated by the fact that not all the neighbors contribute to the current stock trend, we filter out the stocks that are too far away from the current stock (See Section~\ref{eff_emd} for more details). Specifically, for stock $s_i$, we focus on the $k$ nearest neighbors when sorting by the edge weight (which reflects the magnitude of the co-occurrence), where $k$ is a hyper-parameter. Formally, let $S_t(i)$ denote the set of $k$ nearest neighbors of $s_i$ in $G_t$ at time $t$. 

We introduce an attention mechanism~\cite{vaswani2017attention} to infuse the neighbors embedding weighted by an assigned attention value. In this way, we reward the stocks offering more forecasting power by assigning them larger attention values.

\vspace{-3mm}
\begin{equation} 
c^i_t = \sum_{j \in S_t(i)} \alpha_{ij}e_j, 
\label{attention}
\end{equation} 
\vspace{-2mm}
\begin{equation}
    \sum_{j \in S_t(i)} \alpha_{ij} = 1 \textnormal{   for  } j \in S_t(i), 
\vspace{-2mm}
\end{equation}

where $c^i_t$ denotes the fine-tuned stock representation for stock $i$ at time $t$, and $\alpha_{i,j} \in \reals$ is the attention weight on the embedding $e_j$, which is given by 
\vspace{-2mm}
\begin{equation}
    \alpha_{ij} = \frac{exp(f(e_i,e_j))}{\sum_{l \in S(i)}exp(f(e_i,e_l))}.
    \label{update_alpaha}
\end{equation}
The weights define which neighboring stocks are more significant. $f(e_i,e_j)$ measures the compatibility between embeddings $e_i$ and $e_j$, and is parameterized by a feed-forward network with a single hidden layer, which is jointly trained with other parts of the model. We let $f(\cdot,\cdot)$ have the following functional form
\vspace{-1mm}
\begin{equation}
    f(e_i,e_j) =\mathbf{v}_a^{\transpose} \text{tanh}(\mathbf{W_a}[e_i;e_j] +b_a),
\end{equation}
where 
$\mathbf{v}_a$ and $\mathbf{W}_a$ are weight matrices, and $b_a$ is the bias vector~\cite{vaswani2017attention},  obtained during model training via backpropagation.


\section{Leverage heterogeneous data sources}\label{data_source} 

In this section, we show how to integrate heterogeneous data sources for making forecasts, and how we gather different sources of signals.

\vspace{-2mm}
\subsection{Sequential modeling}
Finally, we integrate the stock embedding from \textsc{\modelname} with stock dynamic features into the neural net model to predict the future return.  The stock dynamic input is given by $g^i_t$, which stems from technical factors and news data. We overlay the stock vector $c^i_t$ with $g^i_t$ as an input. Specifically, let $h_t^i$ be the hidden state at time $t$ for stock $i$
\vspace{-2mm}
\begin{equation}
    h_t^i = [c^i_t, g^i_t], 
    \label{concat_input}
\end{equation}
where [$\cdot$, $\cdot$] denotes a direct concatenation.

Keeping in mind that stock trends are highly influenced by a variety of time-series market signals, it is intuitive to take the historical features of a stock as the most influential input to predict its future trend. Therefore, we use Recurrent Neural Networks  \cite{mikolov2010recurrent,schuster1997bidirectional} as the neural net model. LSTM is a variant of the recurrent net, which is capable of learning long-term dependencies.  The final output is given by 
\begin{equation} 
v_{t-T}^i,..., v_{t-1}^i, v_t^i = \text{LSTM}(h_1^i,h_2^i...h_T^i;\theta_l), 
\label{lstm} 
\end{equation} 
where $\theta_l$ denotes the  parameters from LSTM. 

\myparab{Temporal attention layer.}  Since a stock's historical data contributes to its  price trend unequally, we adopt the attention mechanism at the temporal level.  We consider 
\begin{equation}
\begin{matrix}
        r_{t+1}^i = \sum_p \beta_p v_p^i,  \\  \\
\beta_{p} = \frac{exp(f(v_p^i,v_q^i))}{\sum_{q}exp(f(v_p^i,v_q^i))}, 
\end{matrix}
\label{get_r}
\end{equation}
where $\beta_p$ is the attention weight for prior date $p$ indicating the importance of the date. We then compute the weighted sum to incorporate the sequential data and temporal attention. 

Assume we have $m$ trading days and $n$ stocks. We use the mean squared error as the loss function for gradient descent, given by 
\vspace{-1mm}
\begin{equation}
J = \frac{1}{mn}\sum_{i=1}^n\sum_{t=1}^{m}(r_{t+1}^i-\hat{r}_{t+1}^i)^{2}.
\label{loss_main}
\vspace{-1mm}
\end{equation}

Alg.~\ref{alg1} describes our entire algorithmic training pipeline.
Specifically, Lines 2 \& 3 show the procedure to extract long-term relations, and Lines 7-11 show the steps to extract the  evolving relation.
\begin{algorithm}[t!]
\caption{Our Algorithmic Training Pipeline}
\label{alg1}
\begin{algorithmic}[1]
\REQUIRE Online news corpus, technical factors
\ENSURE Prediction on future $\hat{r}_{t+1}^i$ 
\STATE {Variables: Stock embedding matrix $\mE$, stock attention parameters $\alpha_{i,j}$,  LSTM parameters $\theta_l$, temporal attention parameters $\beta_{i}$.}
\STATE Build the global stock co-occurrence matrix $\mX$.
\STATE Use the matrix factorization on $\mX$ with the loss function (Equation~\ref{loss_raw_regu}) to extract static stock embedding.
\REPEAT
\STATE $s_i$ $\gets$ stock $i$ from universe
\FOR{time stamp $t$}
\STATE Build temporal graph $G_t$
\FOR{stock $j$ in  $S_{i}$} 
    \STATE {Obtain the $\alpha_{i,j}$ with Equation~\ref{update_alpaha}} 
\ENDFOR
\STATE Calculate the stock $i$'s final representation $c^i_t$
\STATE Concatenate with the stock dynamic input to obtain final representation $h_t^i$ with Equation~\ref{concat_input}
\STATE Forecast future return $\hat{r}^i_{t+1}$ using Equation~\ref{get_r}.
\ENDFOR
\STATE Calculate prediction loss $J$ by Equation\ref{loss_main}
\STATE Update parameters based on the gradient of $J$  
\UNTIL{convergence}
\label{alg}
\end{algorithmic}
\end{algorithm}
\vspace{-2mm}
\subsection{Gathering difference sources of alphas}

Financial studies have attributed stock movements to three types of market information, i.e., cross-sectional signals,  numerical technical indicators, and  news features. 
To the best of our knowledge, the proposed framework is the first one that fuses technical factors, financial news and stock embedding together for stock predictions.

\myparab{Stock graph: leveraging cross-sectional signals.} Trading on cross-sectional signals (i.e., when Google goes up, Facebook is more likely to go up) is remarkably difficult because we need to examine all possible relations. 
We leverage news articles that mention multiple stocks to detect correlations between stock prices. Detecting co-movements by news appears to be much more effective than existing methods.
Our \textsc{\modelname} learns both long-term and evolving relations. 

\myparab{Technical factors: hand-built features are more effective.} We note that features extracted by deep learning~\cite{doering2017convolutional,lin2017hybrid} are often less effective than features (technical factors) crafted by financial professionals~\cite{colby1988encyclopedia}. Thus, we overlay an LSTM over technical factors, so that we can simultaneously leverage expertise from financial professionals, and also extract serial correlations from deep learning models.  

\myparab{Financail news}
Advancing development of Natural Language Processing techniques has inspired increasing efforts on stock trend prediction by automatically analyzing stock-related articles~\cite{hu2018listening,chiong2018sentiment,akita2016deep}.
We pre-trained \textsc{Word2vec}~\cite{mikolov2013efficient,mikolov2015computing} on the news corpus from the training data set to produce word embeddings. We average all the word vectors in a piece of news to represent the news vector. For a given date $t$ and stock $i$, we compute the daily news vector by extracting the news related to stock $i$ and averaging the news vectors within date $t$. 


\begin{figure*}[t]
    \centering
    \subfigure[]{
    \includegraphics[width=0.35\textwidth]{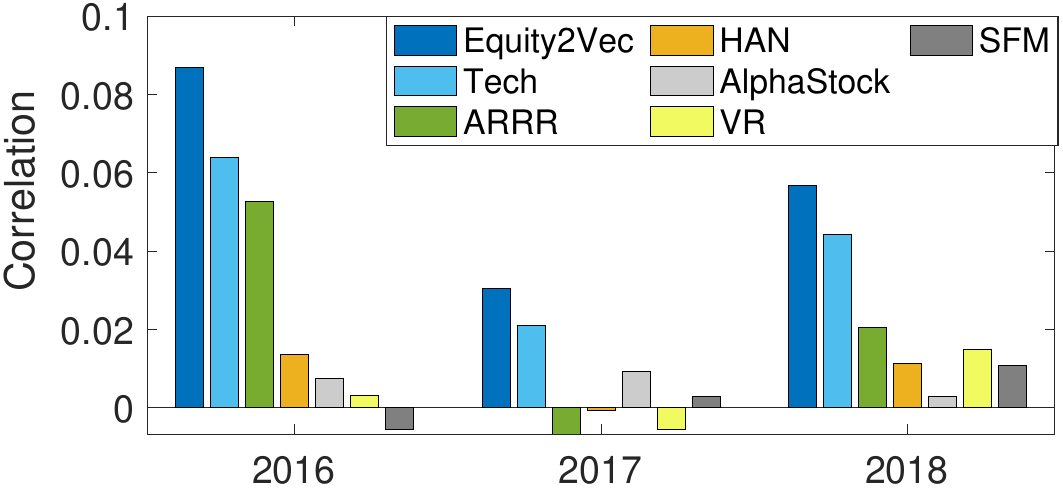}
    \label{fig:corr}}  \hspace{0.1\textwidth}
    \subfigure[]{
    \includegraphics[width=0.35\textwidth]{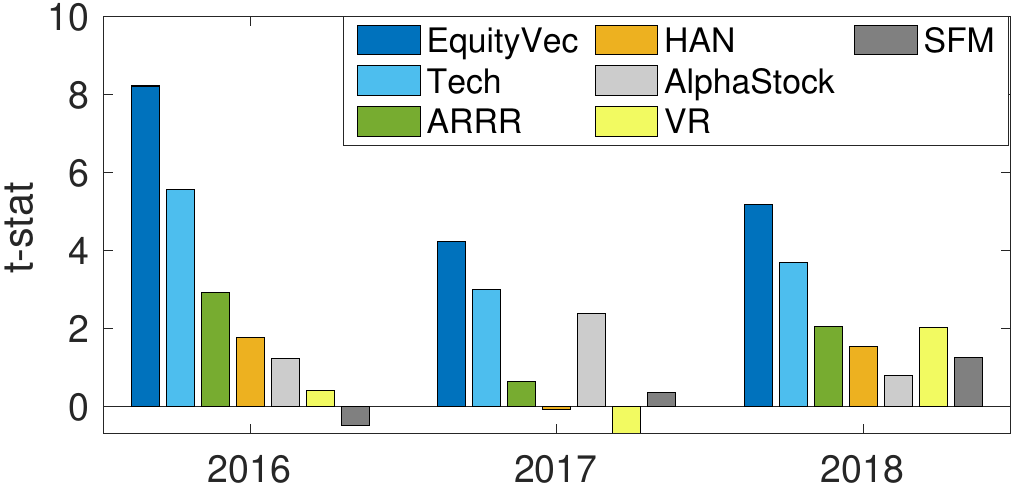}
    \label{fig:tstat} }   
    \vspace{-5mm}
    \caption{Performance comparison  in terms of correlation (a) and $t$-statistic (b) among our \textsc{\modelname}, ARRR, HAN, AlphaStock, VR, and SFM. For both correlation and $t$-statistic, higher scores are better.}
    \label{fig:main_chart}
    \vspace{-4mm}
\end{figure*}
\section{Evaluation Methodology} \label{sec:eval}

We now evaluate the methodology proposed.
We focus on the Chinese market, whose value is the second largest in the world. 

\vspace{-2mm}
\subsection{Data Collection}\label{dataset}
\vspace{-1mm}




\myparab{Chinese Equity Market.}
Our data set consists of daily prices and trading volumes of approximately 3,600 stocks between 2009 and 2018. We consider the universe with all the stocks except for the very illiquid ones. We use open prices (at 9:30 am) to compute the daily returns, and we focus on predicting the next 5-day returns. The last three years of the period  are out-of-sample.

\myparab{Technical Factors. }
We manually build 337 technical factors based on the previous studies~\cite{gu2020empirical,colby1988encyclopedia,kakushadze2016101,amihud2002illiquidity,posner2014economic}. ``Technical factor'' is a broad term encompassing indicators constructed directly from data related to trading activities. Specifically, all these factors are derived from price and dollar volume by mathematical calculation. Table~\ref{top_factors} shows a set of popular technical factors. 
\vspace{-2mm}
\bgroup

\begin{table}[ht!]
\caption{A set of popular technical indicators and the corresponding description.}
\vspace{-4mm}
\Huge
\begin{center}
\resizebox{\linewidth}{!}{
\begin{tabular}{|l|l|}
\hline
Factors & Description \\ \hline
EMA&  Exponential moving average over price or dollar volume. ~\cite{colby1988encyclopedia}  \\ \hline
RSI & The magnitude of one equity's recent price changes. ~\cite{colby1988encyclopedia}\\ \hline
ROC &  Price variation from one period to the next. ~\cite{gencay1998moving}  \\ \hline
Volume Std &  Standard deviation of volume.~\cite{fu2013adopting} \\ \hline
VCR &  Volume cumulative return~\cite{colby1988encyclopedia}\\ \hline
\end{tabular}}
\label{top_factors}
\end{center}
\vspace{-3mm}
\end{table}
\egroup

\myparab{News Dataset.}
We crawled all the financial news between 2009/01/01 and 2018/08/30 from a major Chinese news website Sina\footnote{https://finance.sina.com.cn}. It has a total number of 2.6 million articles. Each article can refer to one or multiple stocks. On average, a piece of news refers to 2.94 stocks. We link each of the collected news articles to a specific stock if the news mentions the stock in the title or content. The timestamps of news published online are usually unreliable (the dates are reliable, but the hour or minute information is usually inaccurate). We use news signals on the next trading day or later to avoid look-ahead issues.

\vspace{-2mm}
\subsection{Experimental settings}\label{setting}
\vspace{-1mm}

\myparab{Training and Testing Data.}
We use three years of data for training, one year of data for validation, and one year for testing. 
The model is re-trained every testing year. For example, the training set starts from Jan 1, 2012 to Dec 31, 2014.
The corresponding validation period is from Jan 15, 2015, to Dec 16, 2015. We use the validation set to tune the hyperparameters and build the model. Then we use the trained model to forecast returns of equity in the same universe from Jan 1, 2016 to Dec 31, 2016, where we set 10 trading days as the ``gap''.   We set a ``gap'' between training and validation periods, and validation periods and testing periods to avoid look-ahead issues. 
The model is then re-trained by using data in the second training period (2013 to 2015) to make forecast on the second testing year.

\myparab{Parameter Setting. }
We use the standard grid search to select the hyper-parameters in our experiments. We build the global co-occurrence matrix in Section~\ref{static} by using all the news before the first day of the testing year. To learn the stocks' embedding, we tune the dimension of stocks' representation within \{32, 64, 128, 256\}. We explore the number of LSTM cell within \{2, 5, 10, 20\}. We greedily search the number of neighbors for the stock graph from no neighbors to all the neighbors.
The sliding window for temporal graph $w$ is tuned within \{2, 5, 10, 20, 60\}.  In addition, we tune the learning rate within \{0.001, 0.01\} with the Adam optimizer~\cite{kingma2014adam}, and set the batch size within \{128, 256\}. 

\myparab{Evaluation Metrics. }
We evaluate our performance in terms of correlation, $t$-statistic, and PnL. 

\mypara{Correlation. }
Unlike the other regression tasks, correlation~\cite{benesty2009pearson} is a preferable metric in stock price prediction compared to MSE since the direction instead of magnitude is more crucial for forecasting return.


\mypara{Significance test ($t$-statistics).}
The use of $t$-statistics estimators~\cite{newey1986simple} can account for the serial and cross-sectional correlations. Recall that $\mr_t \in \reals^n$ is a vector of responses and $\hat \mr_t \in \reals^n$ is the forecast of a model to be evaluated. We examine whether the signals are correlated with the responses, i.e., for each $t$ we run the regression model $\mr_t = \beta_t \hat \mr_t + \epsilon$, and test whether we can reject the null hypothesis that the series $\beta_t = 0$ for all $t$. Note that the noises in the regression model are serially correlated, and thus we use the Newey-West~\cite{newey1986simple} estimator to adjust for serial correlation issues.

\mypara{PnL.}
Profit \& Loss (PnL) is a standard performance measure used in trading. PnL captures the total profit or loss of a portfolio over a specified period. The PnL of all forecasts made on day $t$ is given by 
\vspace{-2mm}
\begin{equation}
\text{PnL} =\frac{1}{n}\sum_i^n  sign(\hat{r}_t^i) * r_t^i ,\quad t = 1,\ldots,m, 
\vspace{-3mm}
\end{equation}

\vspace{-2mm}
\subsection{Baselines for comparison}\label{compare}
\vspace{-1mm}

To test our proposed deep learning framework, 
we compare our model against state-of-art baselines, as described below. For all the baselines, we use the validation data set to configure the hyper-parameters.

\myparab{SFM~\cite{zhang2017stock}.} \textsc{SFM} is designed for stock prediction. It decomposes the hidden states of an LSTM~\cite{rather2015recurrent} network into multiple frequencies by using Discrete Fourier Transform (DFT) in order for the model to capture signals at different horizons.

\myparab{HAN~\cite{hu2018listening}.} This work introduces a so-called hybrid attention technique that translates news into signals. HAN uses Word2Vec to transfer news into vectors, and uses RNN as the  modeling method. 

\myparab{AlphaStock~\cite{wang2019alphastock}.}
AlphaStock integrates deep attention networks reinforcement learning with the optimization of the Sharpe Ratio. For each stock, AlphaStock uses LSTM~\cite{sak2014long} with attention on hidden states to extract the stock representation. Next, it relies on CAAN, a self-attention layer, to capture the inter-relations among stocks. We implement LSTM with basic CAAN, and change the forecast into returns, instead of winning scores. 

\myparab{Vector Autoregression. }
We include a standard linear vector autoregression (VAR)~\cite{negahban2011estimation}. VAR is a typical stock forecast baseline. Formally, it assumes $\mr_{t+1} = f(\mx_t) + \xi_t$,
where  $\mx_t$ denotes the features of all stocks,  and $\mr_{t+1} \triangleq (r_{t+1, 1}, \dots , r_{t+1, n})$ denotes the future return of all stocks in the universe. 

\myparab{ARRR~\cite{wu2019adaptive}. } ARRR is a new regularization technique designed to address the overfitting issue in vector autoregression under the high-dimensional setting. Stock prediction is one of its  applications. Specifically, ARRR involves two SVD steps; the first SVD is for estimating the precision matrix of the features, and the second SVD is for solving the matrix denoising problem.

\vspace{-2mm}
\section{Performance and Discussion}\label{performance}
In this section, we  discuss our overall performance, analyze the effectiveness of $k$-nearest neighbors, the learned stock embedding, and market trading simulation results. 
 
\myparab{Overall Performance on Correlation and $t$-statistic. }
Figure~\ref{fig:corr}  and Figure~\ref{fig:tstat} report the comprehensive analysis on 
all compared methods, for each testing year in terms of both correlation and $t$-statistic. 
The results confirm that our \textsc{\modelname} method consistently outperforms all other baselines, for each testing year and across all metrics.

\begin{figure}[t]
    \centering
    \subfigure[ \textit{ Varying $k$ in testing year 2016.}]{
    \includegraphics[width=0.21\textwidth]{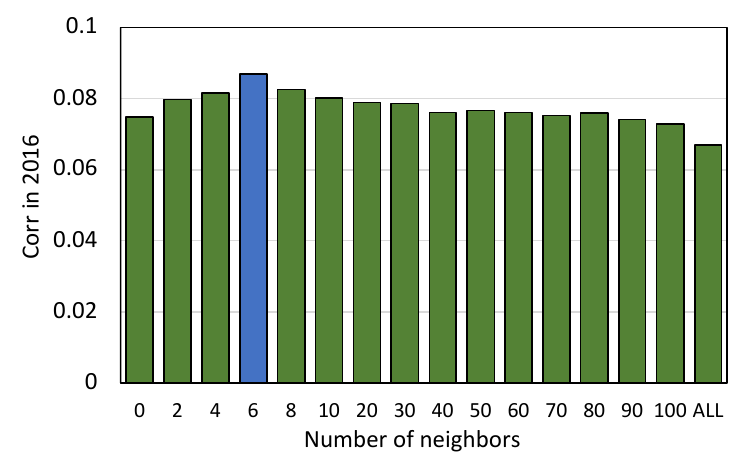}
    \label{fig:k}}  
    \subfigure[ \textit{Different embedding methods.}]{
    \includegraphics[width=0.23\textwidth]{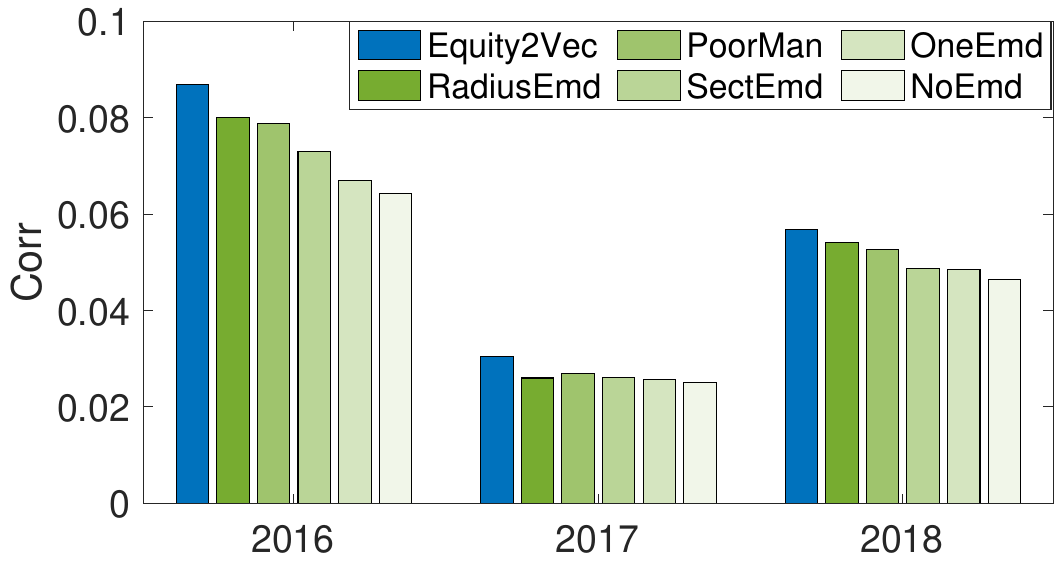}
    \label{emd} }     
    \vspace{-5mm}
    
    \caption{The effects of using different number of neighbors and effects of learned stock representation.}
    \vspace{-7mm}
    \label{neighbors}
\end{figure}

\myparab{Impact of Different Number of Neighbors. }
We investigate the number of neighbors $k$ against the correlation metric. Figure~\ref{fig:k} shows that with the increase of $k$, the out-of-sample correlation first increases and then decreases. This indicates the performance gains from our choices on $k$-nearest neighbors graph in Section ~\ref{dynamic}.

\myparab{Effect of Learned Stock Representation. }\label{eff_emd}
To demonstrate the effects of learned stock representations for stock prediction, we investigate the performance of \textsc{\modelname} by replacing the learned stock representations with the following representations:

\begin{itemize}
    \item NoEmd: Remove the \textsc{\modelname} module. 
    \item PoorMan: Remove the influence from neighbors.
    \item SectEmd: Replace the embedding with the aggregated embedding from the same sector.
    \item OneEmd:  Replace the embedding with the embedding from all the neighbors. 
    \item RadiusEmd: Instead of using $k$-nearest neighbors, we also used the radius to select neighbors and the radius is a hyper-parameter.
\end{itemize}

As shown in Figure.~\ref{emd}, \textsc{\modelname} achieves better performance than the above five baselines. We have the following observations:
(1)  Comparing with NoEmd proves the effectiveness of the information aggregated through stock embedding. (2)  Comparing with PoorMan and OneEmd enables us to confirm the efficacy of learning evolving relations from $k$-nearest neighbors. (3) Comparing with  SectEmd shows that \textsc{\modelname} not just capture merely the sector information. (4) The RadiusEmbd is our variation and also achieved competitive results.

\myparab{Market Trading Simulation. } To further evaluate our method's effectiveness, we conduct a back-testing by simulating the stock trading for three out-of-sample years.
We simulate investments on our signals in two ways:(i) \textit{Long-short} portfolio.
 (ii) \textit{Long-index} portfolio: Long-only minus the market index. We conduct the trading in the daily granularity and select the stocks from the top 20\% strongest forecast signals. 
The position of each stock is proportional to the signal (i.e., the dollar position of $i$-th stock is proportional to our forecast $\hat r^t_i$). The holding period is 5 days. 
\footnote{Short is implementable in the Chinese market only under particular circumstances, e.g., through brokers in Hong Kong under special arrangements. }By allowing short-selling,  we can execute on negative forecasts to understand the overall forecasting quality.
\begin{figure}[t]
    \centering
    \subfigure[ \textit{Long-short PnL}.]{
    \includegraphics[width=0.4\textwidth]{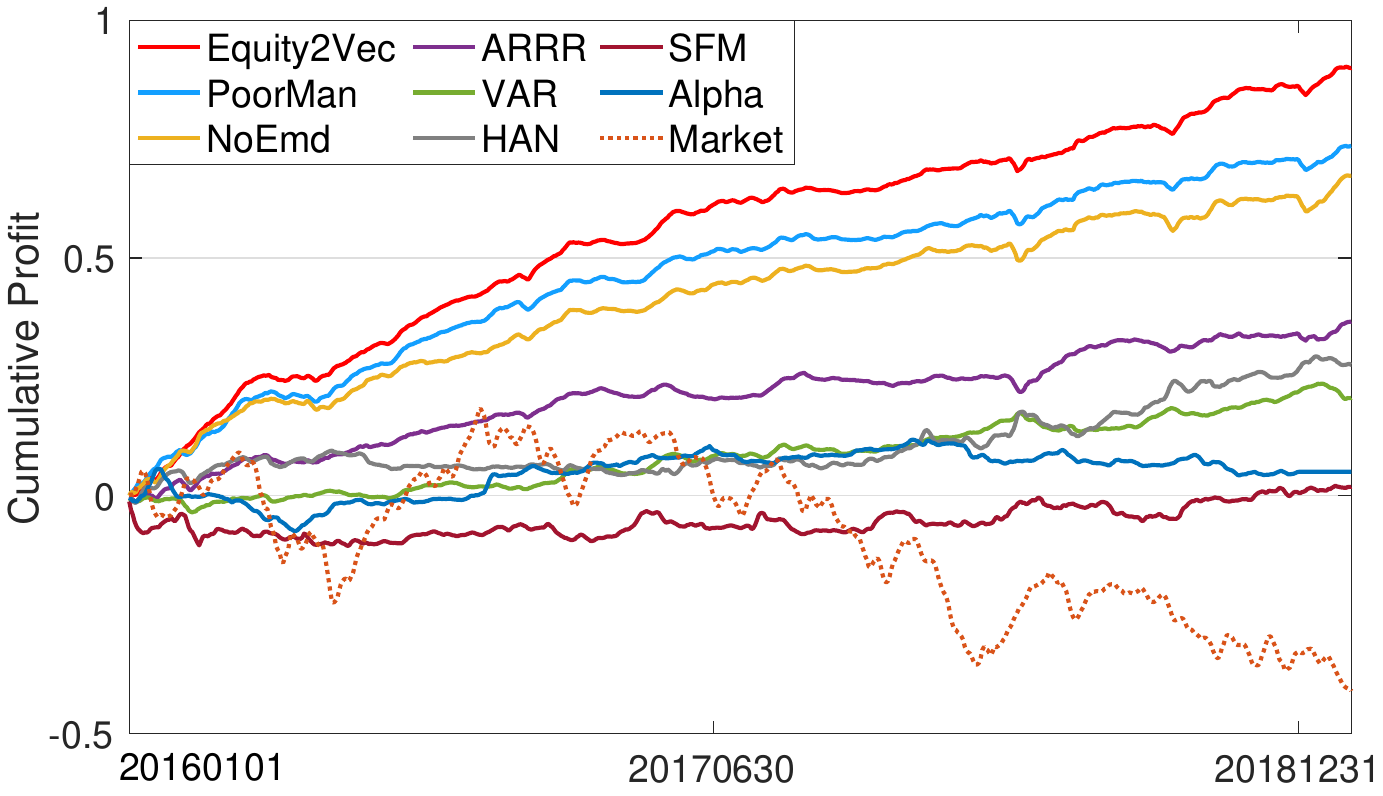}
    \label{fig:quant_ALL}}  \vspace{-0.1mm}
    \subfigure[ \textit{Long-only PnL}.]{
    \includegraphics[width=0.4\textwidth]{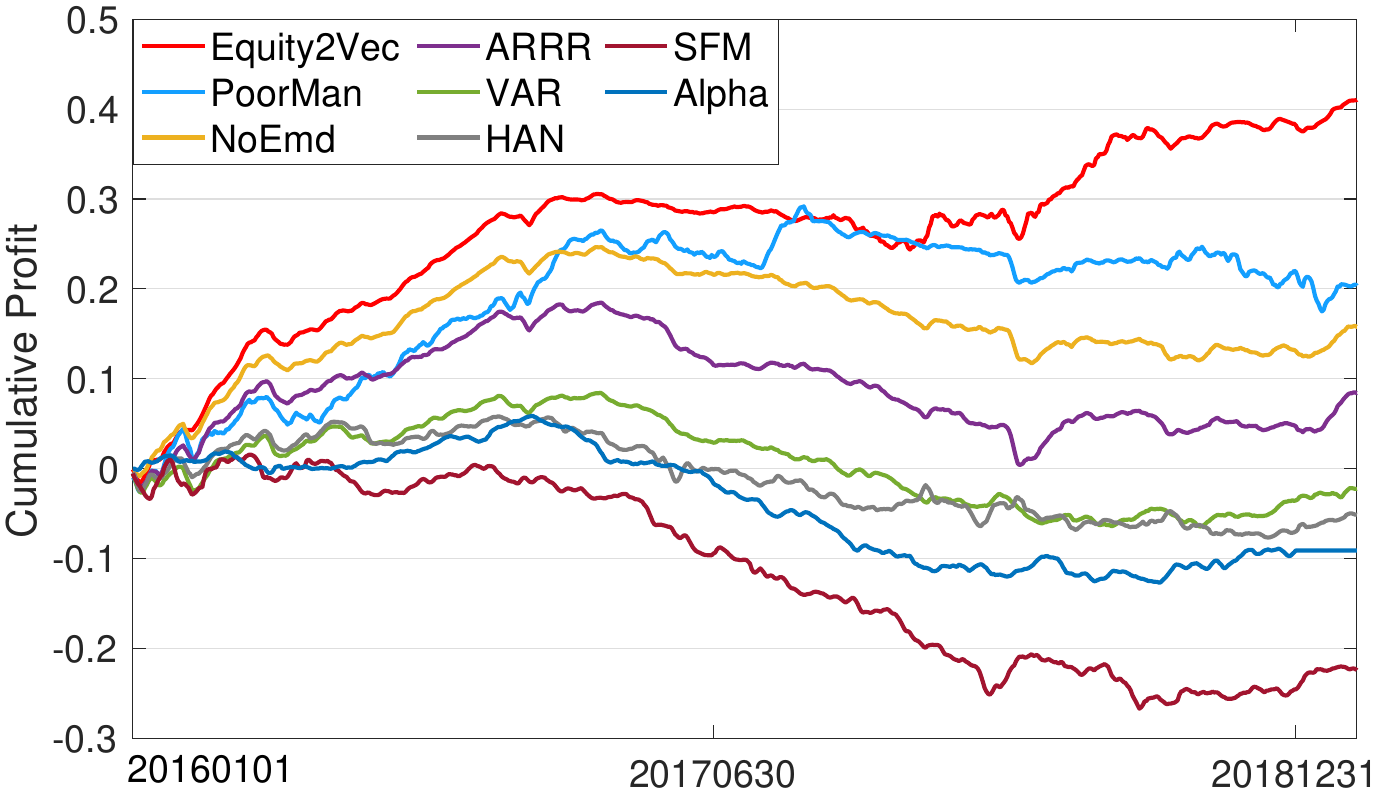}
    \label{fig:quant_LONG} }
    \vspace{-5mm}
    \caption{The cumulative PnL (Profit and Loss) curves of the top quintile portfolio. For example, on any given day, we consider a portfolio with only the top 20\% strongest predictions in magnitude, against future \textit{market excess returns}.
    We simulate the investment on both
     (a) \textit{Long-short portfolio} and (b) \textit{Long-index} portfolio.}
    \label{pnl}
    \vspace{-6mm}
\end{figure}
Figure.~\ref{pnl} shows the cumulative PnL for our approach and baselines.
We can see that our signals are consistently better than other baselines in both  \textit{Long-short} and \textit{Long-index} portfolios, suggesting that our method generates stronger and more robust signals for trading.

\begin{figure}
    \centering
    \includegraphics[scale = 0.4]{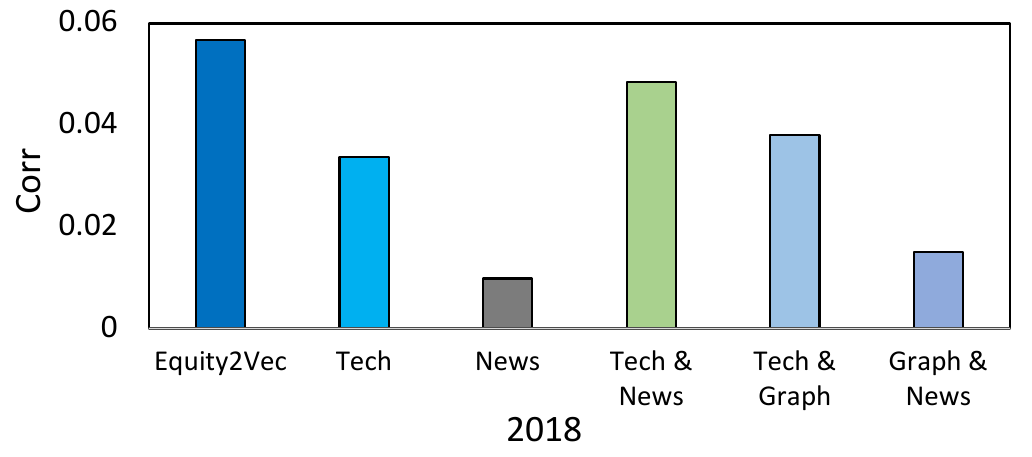}
     \vspace{-4mm}
    \caption{The performance of ablation study.}
     \vspace{-4mm}
    \label{fig:abl}
\end{figure}

\myparab{Ablation study}
To test the effectiveness of different parts of the \textsc{\modelname}, we conduct an ablation study that excludes components and measures the out-of-sample correlation in the year 2018. The results are shown in Figure~\ref{fig:abl}.
We draw two main conclusions from these results. \emph{(i) Each component has a substantial impact on the performance:} the results validate the inclusion of each component in our method.
\emph{(ii) Effect of cross-categorical learning:} the model using graph achieves higher performance than the baselines without it, thus illustrating its ability to capture and leverage latent stock interactions. 

\vspace{-2mm}
\section{Interpretation Analysis}\label{interpretation}
\vspace{-1mm}

In this section, we assess the interpretability for stock relationships, temporal weights in LSTM, and news as follows.

\begin{figure}
    \centering
    \includegraphics[scale = 0.22]{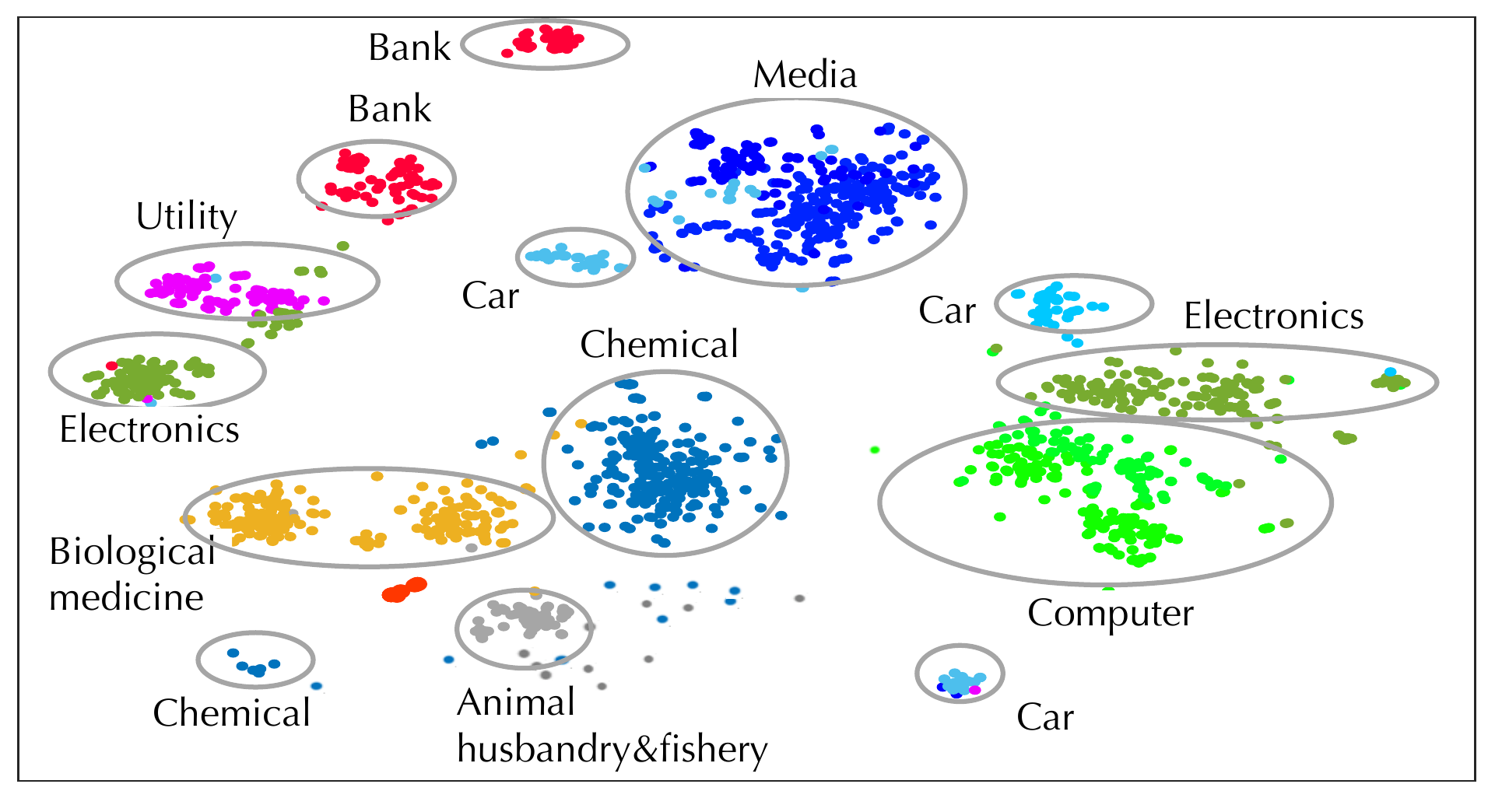}
    \caption{t-SNE of final stock representations (colors code industry sectors).}
    \vspace{-4mm}
    \label{fig:tsne_stock}
\end{figure}

\myparab{Visualizing Learned Stock Embedding.} As depicted in Figure~\ref{fig:tsne_stock}, we use t-SNE~\cite{maaten2008visualizing} on our final stock representation to assess the interpretation of the universe of stocks. Each dot represents a stock, and the color denotes the largest sector from Barra~\cite{fabozzi2011theory} associated with the given stock,  while the text annotations represent the detailed stock categories. It shows that our \textsc{\modelname} learns the interpretable stock representations that are well aligned with the Barra sectors.

\begin{figure}
    \centering
    \includegraphics[scale = 0.36]{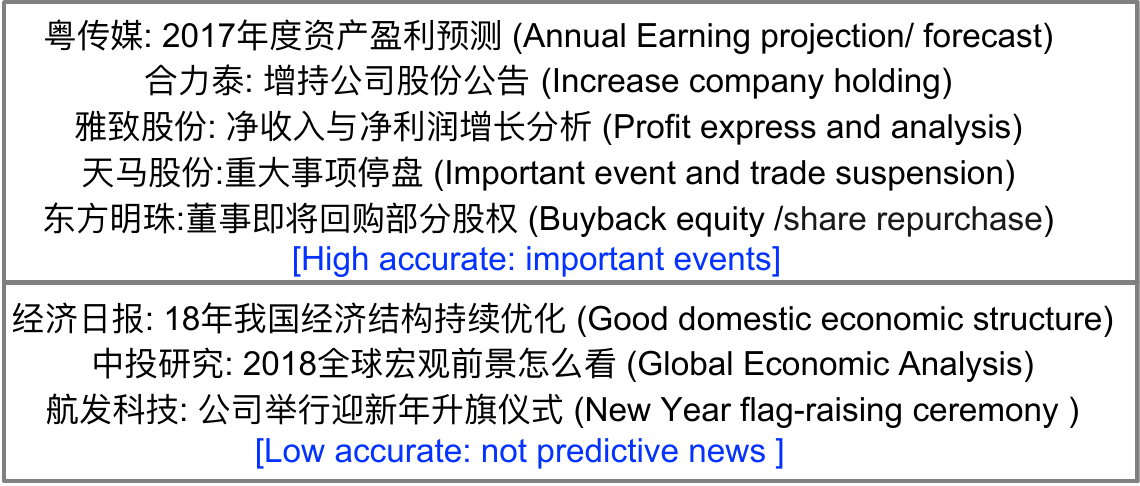}
    \vspace{-4mm}
    \caption{The demonstrative news from high accuracy and low accuracy performance.}
    \vspace{-5mm}
    \label{fig:news_class}
\end{figure}

\myparab{News Interpretation. }
To understand the news predictive ability, we track back the news from two groups in the test dataset.  We focus the samples within the smallest $5\%$ and the samples within the largest $5\%$ errors based on Equation~\ref{loss_main}. We extract the corresponding news and show the detailed results in Figure~\ref{fig:news_class}. We focus on the demonstrative news from these two groups of samples. One can see that the news in the high accuracy group contains significant events with predictive ability, while the news in the low accuracy group mainly has no apparent influence on the stock forecast.

\myparab{Temporal Attention Explanation. }
Next, we show the overall attention weights from the testing data set in Table~\ref{tlb:attention_weight} when we use the past 5 days' historical data. The hyperparameter 5 is set by the performance in the validation dataset. The recent days have larger weights indicating the recent days play more significant roles in the prediction.

\vspace{-3mm}
\bgroup
\begin{table}[H]
\Huge
\begin{center}
\resizebox{0.8\linewidth}{!}{\begin{tabular}{|l|l|l|l|l|l|}
\hline
        & -5day & -4day & -3day & -2day & -1day \\ \hline
Weights &  0.0055   &  0.0265   &  0.1662  &    0.3064  &   0.4954  \\ \hline
\end{tabular}}
\captionsetup{font=small}
\caption{The temporal overall attention weights.}
\vspace{-0.7cm}
\label{tlb:attention_weight}
\end{center}
\end{table}
\egroup

\vspace{-2mm}
\section{Related work}\label{related}
\vspace{-1mm}

Forecasting future prices of equities has been extensively studied~\cite{wuthrich1998daily,campbell1997econometrics,grinold2000active,bali2016empirical,kelly2019characteristics,harvey2016and,cakici2017cross,kozak2020shrinking,fama2016dissecting,heaton2017deep,gu2018empirical,chen2019investment,wang2019alphastock,feng2019temporal,huang2019shrinking,friedman2001elements,han2018firm,de2018advances,feng2018deep,chen2019deep,gu2019autoencoder,chen2019investment,ding2015deep,zhang2017stock,hu2018listening,ke2019deepgbm,li2019individualized}. A wide range of practical machine learning~\cite{wang2019alphastock,8665577,hu2018listening,liu2019near,cluster21-weighttransfer,wu2015early} techniques such as deep learning, boosting tree, and non-parametric methods are introduced to forecast asset prices.
Most have examined how the fundamental characteristics, such as the book/market ratio, capital, and momentum of the issuing firm could affect the price of the asset over the long term, or how various events or macroeconomic factors could cause price changes~\cite{das1998earnings,pritamani2001return,kelly2019characteristics}. The studies have tended to use  ``low-frequency" models because the response horizon is in the order of months (e.g., how a stock's price changes after 3 months). Linear statistical models need to be used because the data sets are remarkably small, e.g., each asset is associated with only a few data points (e.g., each month corresponds to a data point). 
``Mid-frequency'' models with one or multiple days of forecasting horizon had been studied more extensively in the private sectors. It has been observed that trading activities (e.g., price/volume changes) have a heavier impact on an equity's short term price movement than fundamental factors so significant effort was devoted to understanding the prediction powers of ``technical'' factors such as volatility or trading volume changes~\cite{wang2019alphastock,chen2019investment,feng2019temporal} for empirical asset pricing. Because we can collect a more reasonable amount of data (e.g., one trading day results in one data point), using machine learning techniques to extract interactions becomes possible. Recent years have witnessed explosive development of mid-frequency ML models from both finance and computer science. 

``High-frequency'' models investigate price changes over an even shorter time horizon (e.g., next 10-second return). The price changes in this horizon are usually driven by the market microstructure (the dynamics of the bid-ask queues). Here, the modeling challenge is more on the computational side. It is often remarkably to fit an ML model properly with an abundant amount of data. So a significant fraction of effort focuses on speeding up the system, which resembles typical system and architecture research~\cite{wu2020rosella,liu2020gpus,ibrahim2019analyzing,liu2018architectural,lin2018gpu,liu2016lightweight}.

Our work examines the mid-frequency models and focus on recent work using machine learning to forecast asset prices.

\myparab{Linear Model. }
Empirical asset pricing (e.g., estimating the ``true price'' or forecasting the future price) is an important area in Finance~\cite{bali2016empirical}. Linear regression has been a dominating methodology to forecast equity returns, especially for intraday tradings~\cite{harvey2016and,cakici2017cross,kozak2020shrinking,fama2016dissecting}. Because these linear models usually use a large number of features, regularizations are usually needed~\cite{huang2019shrinking,kelly2019characteristics,negahban2011estimation,wu2019adaptive}. For example, recently researchers examined regularization for ``ultra-high dimensional'' setting, in which the number of features could be significantly larger than the number of observations~\cite{wu2019adaptive}.

\myparab{Deep Learning.} There are two major approaches to forecast equity returns. \emph{Approach 1. ANN as a blackbox for standard ``factors.''} First, ``factors'' that are known to be correlated with returns are constructed. These factors can be viewed as features constructed by financial experts. Second, the factors are fed into standard ANN black boxes so that non-linear models are learned
(see e.g.,~\cite{heaton2017deep,gu2018empirical} and references therein). Little effort is made to optimize ANN's architecture or algorithm. \emph{Approach 2. Forecasting the price time-series.} This approach views the price, trading volume, and other statistics representing trading activities as time series and designs specialized deep learning models to extract signals from the time series. Little feature engineering is done for these models. See e.g.,~\cite{feng2018deep,rather2015recurrent,doering2017convolutional,lin2017hybrid,sezer2020financial,lim2019enhancing}. Approach 1 represents the line of thought that feature engineering is critical in building machine learning models, whereas Approach 2 represents the mindset that deep learning can automatically extract features so effort on feature engineering should be avoided. The co-movement effect is often ignored by the previous studies. Only a few works on stock predictions have explored this effect~\cite{chen2019investment,wang2019alphastock,feng2019temporal}. However,
\cite{chen2019investment} relies on non-public dataset and learns the relations in a static way. \cite{feng2019temporal} consider only consider the relation to a particular type (sector and supply chain) and ignore the other relations, such as stocks affected by the same event. \cite{wang2019alphastock} has unsatisfying performance (even in their own reports on experiments) and only consider the  co-movements of historical prices.

\myparab{News.} NLP-based techniques~\cite{wu2020bats} are developed to correlate news with the movement of stock prices. Earlier works use matrix factorization approaches (see e.g., ~\cite{ming2014stock,sun2016trade}) whereas more recent approaches use deep learning methods~\cite{ding2015deep,hu2018listening,chiong2018sentiment,akita2016deep}. These methods exclusively use the news to predict equity returns, and they do not consider any other ``factors'' that can impact the stock prices. 

\myparab{Factor Model (Cross-sectional Returns). } The movement of two or more stocks usually can be explained by a small subset of factors. For example, $\mathrm{Facebook}$ and $\mathrm{Google}$ often co-move because their return can be explained by the technical factors. The so-called ``factor model'' (e.g.,~\cite{Fama,stock2011dynamic,zhang2017stock,wu2019adaptive}) can effectively capture the co-movement of prices but these methods usually rely on PCA/SVD techniques and are not computationally scalable. 


\myparab{Comparison.} \emph{1. Comparing to existing linear models.} Our method is more effective at extract non-linear signals. \emph{2. Comparing to existing DL models.} We find that we need both careful feature engineering and optimizing DL techniques to use technical factors in the most effective manner, moreover, we do not restrict the stock relation into a particular type and learn the evolving relations. 
\emph{3. Comparing with News/NLP-based techniques.} We do not exclusively rely on the news. Instead, we explicitly model the interaction between news and other factors so that our model avoids low quality signals (e.g., news-based signals could be essentially trading momentum), and \emph{4. Comparing with factor models.} A key innovation of our model is the introduction of \textsc{\modelname} component. This component models the interaction between stocks and circumvents SVD computation on large matrices. 

\vspace{-2mm}
\section{Conclusion and future work}\label{con}
\vspace{-1mm}
This paper presents a novel approach to answer two research questions. 
\emph{(i)} How can we interpret the relationship between stocks? 
\emph{(ii)} How can we leverage heterogeneous data sources to extract high-quality forecasting power? 
Through extensive evaluation against the state-of-art baselines, we confirm that our method achieves superior performance. Meanwhile, the results from different trading simulators demonstrate that we can effectively monetize the signals. In addition, we interpret the stock relationships highlighting they align well with the sectors defined by commercial risk models, extract important technical factors, and explain what kind of news has more predictive power.

We identify several potential future directions. First, it is worth exploring more effective features from social media such as financial discussion forums. As individual investors often engage in insightful discussions on finance topics and stock movements, the large volume of such discussions could indicate potential upcoming major events. Second, 
the proposed \textsc{\modelname} can generalized to other problems, such as mining the relations between futures.

\section{Acknowledgement}

We thank anonymous reviewers for helpful comments and suggestions. Christopher G. Brinton was supported in part by the Charles Koch Foundation. Qiong Wu, Zheng Zhang and Zhenming Liu are supported by NSF grants NSF-2008557, NSF-1835821, and NSF-1755769.
Mihai Cucuringu and Andrea Pizzoferrato acknowledge support from The Alan Turing Institute EPSRC grant EP/N510129/1.  Andrea Pizzoferrato also acknowledges support from the National Group of Mathematical Physics (GNFM-INdAM), and by the EPSRC grant EP/P002625/1. The authors acknowledge William \& Mary Research Computing for providing computational resources and technical support that have contributed to the results reported within this paper.

\bibliographystyle{ACM-Reference-Format}
\bibliography{acmart}


\end{document}